\documentclass[10pt,twocolumn,letterpaper]{article}

\usepackage{cvpr}
\usepackage{times}
\usepackage{epsfig}
\usepackage{graphicx}
\usepackage{amsmath}
\usepackage{amssymb}
\usepackage{}
\usepackage{capt-of,etoolbox}

\newcommand\minisection[1]{\vspace{2mm}\noindent \textbf{#1}}

\usepackage[hyphens]{url}
\usepackage[pagebackref=true,breaklinks=true,letterpaper=true,colorlinks,bookmarks=false]{hyperref}
 \cvprfinalcopy 

\def\cvprPaperID{2275} 

\ifcvprfinal\fi
\makeatletter
\patchcmd\@maketitle\null{{\myfigure{}\par}}{}{}
\makeatother

\begin{document}

\newcommand\myfigure{%
\centering
    \includegraphics[width=0.95\linewidth]{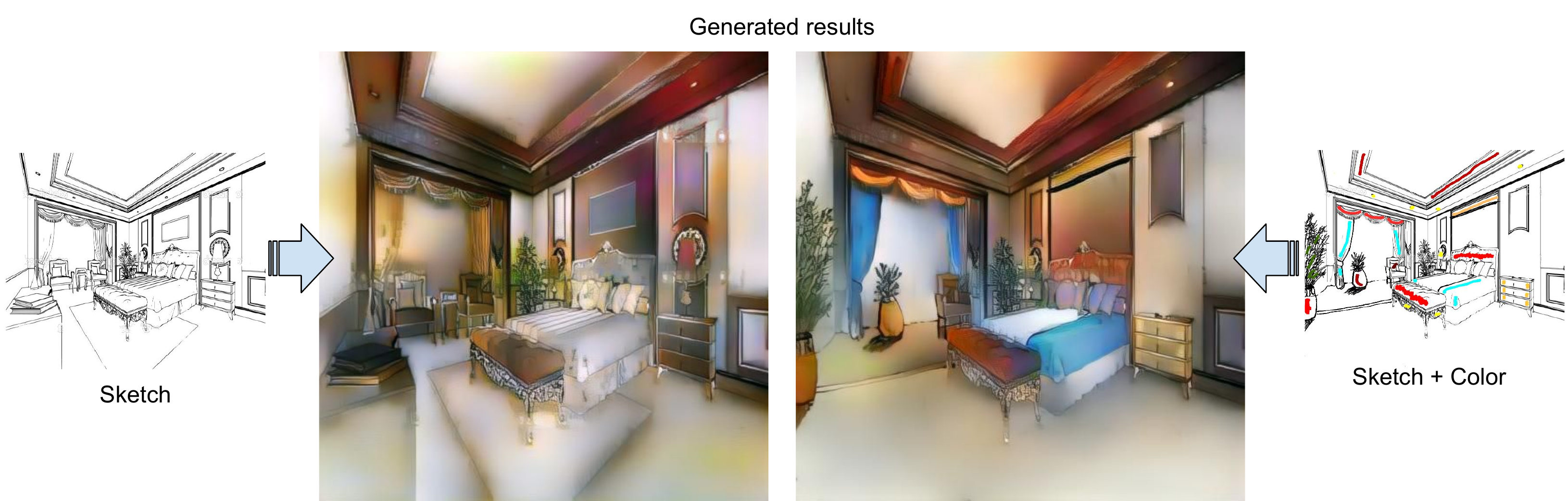}
\captionof{figure}{A user can sketch and scribble colors to control deep image synthesis. On the left is an image generated from a hand drawn sketch. On the right several objects have been deleted from the sketch, a vase has been added, and the color of various scene elements has been constrained by sparse color strokes. For best resolution and additional results, see http://scribbler.eye.gatech.edu}
\label{fig:teaser}
}
\title{Scribbler: Controlling Deep Image Synthesis with Sketch and Color}
\author{Patsorn Sangkloy\textsuperscript{1}, \quad Jingwan Lu\textsuperscript{2}, \quad Chen Fang\textsuperscript{2}, \quad Fisher Yu\textsuperscript{3}, \quad James Hays\textsuperscript{1}\\
\\
\textsuperscript{1}Georgia Institute of Technology \quad \textsuperscript{2}Adobe Research \quad \textsuperscript{3}Princeton University}

\maketitle
\begin{abstract} 
Recently, there have been several promising methods to generate realistic imagery from deep convolutional networks. These methods sidestep the traditional computer graphics rendering pipeline and instead generate imagery at the pixel level by learning from large collections of photos (e.g. faces or bedrooms). However, these methods are of limited utility because it is difficult for a user to control what the network produces. In this paper, we propose a deep adversarial image synthesis architecture that is conditioned on sketched boundaries and sparse color strokes to generate realistic cars, bedrooms, or faces. We demonstrate a sketch based image synthesis system which allows users to ‘scribble’ over the sketch to indicate preferred color for objects. Our network can then generate convincing images that satisfy both the color and the sketch constraints of user. The network is feed-forward which allows users to see the effect of their edits in real time. We compare to recent work on sketch to image synthesis and show that our approach can generate more realistic, more diverse, and more controllable outputs. The architecture is also effective at user-guided colorization of grayscale images. 

\end{abstract}

\section{Introduction}

Recently, numerous image synthesis methods built on neural networks have emerged~\cite{SalHinton07,Lee:2009,goodfellow2014generative,radford2015unsupervised,kingma2013auto,Gregor2015DRAWAR}. These methods can generate detailed and diverse (if not quite photorealistic) images in many domains. However, it is still unclear how to \emph{control} these powerful new tools. How can we enable everyday users (non-artists) to harness the power of deep image synthesis methods and produce realistic imagery? Several recent methods have explored controllable deep synthesis~\cite{Dosovitskiy_2015_CVPR,Attribute2Image,zhou2016view,guccluturk2016sketchinv,zhu2016manifold,pix2pix2016,van2016conditional} and we focus on two complementary forms of control -- sketches and color strokes. Sketches are a compelling form of control because anyone can draw (potentially very badly) and because it is easy to edit sketches, e.g. to remove or add objects, whereas the equivalent operations in the image domain require artistic expertise. Color is a compelling form of control because many sketches or grayscale scenes are fundamentally ambiguous with respect to color~\cite{zhang2016colorful}, but it is easy for a user to intervene, e.g. to scribble that drapes should be blue and the valance should be red (Figure~\ref{fig:teaser}). Both forms of control are relatively sparse and require a deep network to synthesize image detail beyond what is contained in the input. The deep network must also implicitly learn a significant amount of high-level image understanding, e.g. what colors are allowable for particular objects, the boundaries of objects such that color does not bleed beyond a single semantic region, and the appropriate high frequency textures for different scene elements.

We propose a deep adversarial (GAN) image synthesis architecture trained to generate realistic images from sparse and simple sketched boundaries and color strokes. We train our network on a diverse set of synthetic sketches optionally augmented with randomly sampled color strokes. The network learns to recover the color and detail lost to the sketching process and to extrapolate the sparse color indications to semantic scene elements. We show qualitative results of image synthesis in three domains -- faces, cars, and bedrooms. We test on synthetic sketches as well as imperfect hand-drawn sketches. 

Our approach is similar to Sketch Inversion~\cite{guccluturk2016sketchinv}, which also generates images from sketches, although we show the benefit of adversarial training, introduce color control signals, demonstrate results on image domains beyond faces, and demonstrate that users can perform simple edits to sketches to control the synthesis. Our control signals are most similar to Zhu et al.~\cite{zhu2016manifold} -- they also demonstrate that GANs can be constrained by sketch and color strokes. However, our architecture is a feed-forward mapping from sketch and color to images while Zhu et al. perform an optimization to map user sketches into the latent GAN space in order to find the most similar image on the natural image manifold (as understood by the GAN). Their approach does not see user inputs at training time and thus cannot learn the complex mapping between user inputs and desired image outputs. Their method is also significantly slower because it is not a strictly feed-forward process and this hinders interactive image editing. The concurrent work of Isola et al.~\cite{pix2pix2016} significantly overlaps with our own. Both approaches use conditional GANs for the sketch to photo as well as grayscale to color synthesis tasks, although they do not focus on user control of the synthesis.

The contributions of this paper include:
\begin{itemize}
  \item First and foremost, we are the first to demonstrate an adversarial deep architecture that can learn to generate realistic images from imperfect sketches with sparse color 'scribbles'. Our feed-forward architecture is fast and interactive. 
  \item We improve the quality of sketch-to-image synthesis compared to existing work~\cite{guccluturk2016sketchinv}. We produce higher resolution, more diverse images spanning more image domains (bedrooms and cars in addition to faces).
  \item We show that our method can generate realistic images from diverse sketch styles, including imperfect human sketches or edits of synthetic sketches. We achieve this generality by augmenting our training data with multiple sketch styles.
  \item Finally, we demonstrate that our adversarial architecture is also promising for image colorization. We show encouraging results for grayscale to RGB conversion and introduce controllable colorization using the same sparse color strokes used with sketches.  
\end{itemize}

\section{Related Work}

Synthesizing images by learning from image collections is a long standing interest of the computer graphics and computer vision communities. Previously, the most successful methods tended to be non-parametric approaches which found clever ways to reuse existing image fragments ~\cite{McMillan:1995,EfrosLeung,SceneCompletion,chen2009sketch2photo,barnes2009patchmatch}.

In the last few years, \textit{parametric} models built on deep convolutional networks have shown promising results~\cite{goodfellow2014generative,Dosovitskiy_2015_CVPR,radford2015unsupervised,kingma2013auto,Gregor2015DRAWAR}. While deep image synthesis methods cannot yet create realistic, high-resolution images they have an implicit ability to generalize that is difficult for data-driven non-parametric methods (e.g. the ability to hallucinate unseen viewpoints of particular chairs based on the appearance changes of other chairs~\cite{Dosovitskiy_2015_CVPR}). Because our visual world is both enormously complex (with appearance depending on viewpoints, materials, attributes, object identity, lighting, etc.) \emph{and} heavy-tailed, non-parametric methods are limited even in the ``big data'' era. But deep image synthesis methods might implicitly factorize our visual world and thus generalize to situations beyond the training examples.

A common approach to deep image synthesis is to learn a low dimensional latent representation that can later be used to reconstruct an image, e.g. with Variational Autoencoders (VAEs)~\cite{kingma2013auto} or Generative Adversarial Networks (GANs)~\cite{goodfellow2014generative}. In general, deep image synthesis can be conditioned on any input vector ~\cite{van2016conditional}, such as attributes~\cite{Attribute2Image}, 3d viewpoint parameters and object identity~\cite{Dosovitskiy_2015_CVPR}, image and desired viewpoint~\cite{zhou2016view}, or grayscale image~\cite{zhang2016colorful,IizukaSIGGRAPH2016letbecolor,larsson2016learning} .  

\minisection{Generative Adversarial Networks (GANs)}
Among the most promising deep image synthesis techniques are Generative Adversarial Networks (GANs)~\cite{goodfellow2014generative,radford2015unsupervised} in which a \textit{generative} network attempts to fool a simultaneously trained \textit{discriminator} network that classifies images as real or synthetic. The discriminator discourages the network from producing obviously fake images. In particular, straightforward regression loss for image synthesis often leads to `conservative' networks which produce blurry and desaturated outputs which are close to the mean of the data yet perceptually unrealistic. After training, the generator network is able to produce diverse images from a low dimensional latent input space. Although optimizing in this latent space can be used to 'walk' the natural image manifold (e.g. for image editing~\cite{Brock2016neural_editing,zhu2016manifold} or network visualization \cite{nguyen2016synthesizing,nguyen2016ppgn}), the space itself is not semantically well organized -- the particular dimensions of the latent vector do not correspond to semantic attributes although mapping them to an intermediate structure image~\cite{wang2016generative} can give us more insight.

\minisection{Conditional GANs}
Instead of synthesizing images from latent vectors, several works explore \emph{conditional} GANs where the generator is conditioned on more meaningful inputs such as text ~\cite{reed2016text2image,reed2016learning}, low resolution images (super-resolution)~\cite{ledig2016SRGAN,sonderby2016amortised_sr_adv}, or incomplete images (inpainting)~\cite{pathakCVPR16context,oord2016pixel,yeh2016semantic}. Conditional GANs have also been used to transform images into different domains such as a product images~\cite{yoo2016pixel} or different artistic styles~\cite{li2016precomputed}. Conditional GANs can also condition the \emph{discriminator} on particular inputs, e.g. Reed et al.~\cite{reed2016text2image} condition both the generator and discriminator on an embedding of input text. This effectively makes the discriminator more powerful. In this paper, only our generator is conditioned on input sketches and color strokes leaving the discriminator to discern real vs fake and not to evaluate the appropriateness of an output given the particular input.

\minisection{Controlling deep image synthesis}
Several recent works share our motivation of adding user editable \textit{control} to deep image generation.  Examples of control signals include 3d pose of objects~\cite{Dosovitskiy_2015_CVPR}, natural language~\cite{reed2016text2image}, semantic attributes~\cite{Attribute2Image}, semantic segmentation ~\cite{champandard2016semantic}, and object keypoints and bounding box~\cite{reed2016learning}.

The artistic style transfer approach of Gatys et al.~\cite{gatys2016image} could also be considered a mechanism to control deep image synthesis. Their method does not `learn' transformations end-to-end but instead uses a pre-trained network and optimizes for output images which have deep network feature \emph{activations} (content) similar to one input image and deep network feature \emph{correlations} (style) similar to another input image. The approach does not perform well for transformations which requires the synthesis of realistic detail (e.g. trying to preserve the `content' of a sketch and the `style' of a photograph).

The most similar previous deep image synthesis approach in terms of \emph{control} is Zhu et al.~\cite{zhu2016manifold} which optimizes for an image that is similar to an input sketch (potentially with color strokes) that lies on a learned natural image manifold. However, identifying a matching image within this manifold that is similar content-wise to the sketch can be challenging when the sketch and the image are significantly different. For the `sketch brush' in~\cite{zhu2016manifold}, they get around this by optimizing for image with the same edges as user sketch that also lies within a natural image manifold as approximated by a pre-trained GAN. However, image edges are not necessarily a good proxy for human sketched strokes~\cite{sketchy2016} and their method has no capacity to \emph{learn} the mapping between user inputs and desired outputs. In contrast, our method enables control via sketch and color strokes in a unified framework learned end-to-end. Sketch Inversion~\cite{guccluturk2016sketchinv} is also closely related to our work although they do not address color control. We compare our sketch-to-photo results with Sketch Inversion.
 
\minisection{Controllable Colorization} Our color control strokes are inspired by Colorization Using Optimization~\cite{levin2004colorization} which interpolates sparse color strokes such that color changes tend to happen at intensity boundaries. The algorithm does not learn the association between objects and colors and thus can only interpolate user provided colors (e.g. a tree in the background of a scene will not be green if the user only marked foreground objects). The algorithm also does not learn the spatial extent of objects, and thus colors might `snap' to spurious boundaries or bleed over weak intensity edges that are none-the-less salient boundaries. Our deep network learns object color tendencies and object extent and thus can cleanly color objects either with no color strokes or with color strokes on a subset of scene elements (Figure~\ref{fig:teaser}). Similar control strokes have been applied to sketch and manga imagery~\cite{qu-2006-manga,sykora2009lazybrush}, but the results remain non-photorealistic and lack lighting and shading.
 
We are unaware of sparse scribbles being used as input constraints to deep generative networks, although ScribbleSup~\cite{ScribbleSup} uses sparse scribbles to \emph{supervise} the output of semantic segmentation networks. The scribbles are training data and there is no user control at test time.
 
\minisection{Concurrent work} Concurrent to our work, the `pix2pix' method of Isola et al.~\cite{pix2pix2016} also uses conditional GANs for sketch to photo and grayscale to color synthesis. Additionally, they explore several other interesting image-to-image `translation' tasks. Unlike our approach, they use a ``U-Net'' architecture~\cite{UNet} which allows later layers of the network to be conditioned on early layers where more spatial information is preserved. They condition both their generator and discriminator on the input whereas we condition only the generator. Their results are high quality and they are able to synthesize shoes and handbags from coarse sketches~\cite{Eitz2012} even though their training data was simple image edges. In contrast, we take care to train on a diversity of synthetic sketch styles. The most significant difference between our works is that we introduce sparse color control strokes and demonstrate how to train a network so that it learns to intelligently interpolate such control signals, whereas Isola et al.~\cite{pix2pix2016} does not emphasize controllable synthesis.

\section{Overview}
In this paper, we explore adding direct and fine-grained user controls to generative neural networks. We propose a generic feed-forward network that can be trained end-to-end to directly transform users' control signals, for example a hand-drawn sketch and color strokes, to a high-res photo with realistic textural details. 

Our proposed network is essentially a deep generative model that is conditioned on control signals. The network learns a transformation from control signal to the pixel domain. It learns to fill in missing details and colors in a realistic way. 
Section~\ref{sect:network} discusses the network structure that is shared by all applications presented in the paper. Section~\ref{sec:objective_function} introduces the objective functions, in particular the combination of content loss and adversarial loss, which encourages the result to be photo-realistic while satisfying user's fine-grained control. Section~\ref{sec:sketch2photo} and~\ref{sec:colorization} show how to enforce two different user controls in the proposed framework -- using hand-drawn sketches to determine the gist or shape of the contents and using sparse color strokes to propagate colors to semantic regions. Section~\ref{sec:apps} applies the proposed framework in several interactive applications.

\begin{figure}[ht]
\begin{center}
\includegraphics[width=1\linewidth]{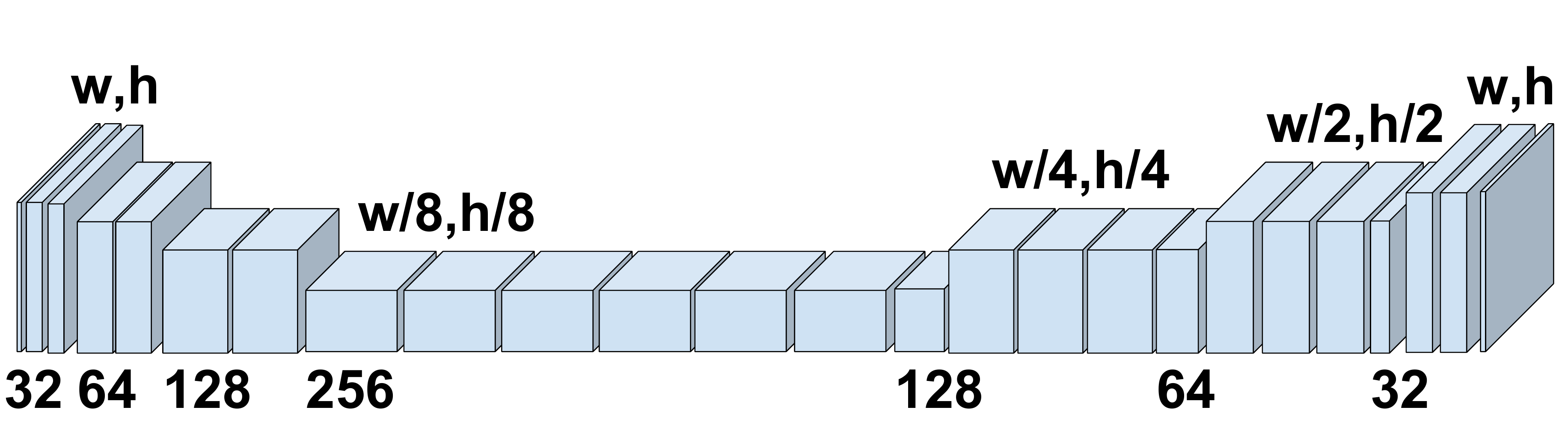}
\end{center}
   \caption{Network Architecture. Following the encoder-decoder design, we use three downsampling steps, seven residual blocks at the bottleneck resolution and three upsampling steps. Residual blocks use stride 1. Downsampling uses convolutions with stride 2. Upsampling uses bilinear upsampling followed by residual blocks.}
\label{fig:network_diag}
\end{figure}

\subsection{Network Architecture}
\label{sect:network}
We design a feed-forward neural network that takes an image as input and generates a photo of the same resolution as output. When generating an image conditioned on a high dimensional input in the same domain (i.e. from image to image), typically an encoder-decoder type of network architecture is adopted, for example in sketch inversion \cite{guccluturk2016sketchinv}, image colorization \cite{zhang2016colorful,IizukaSIGGRAPH2016letbecolor}, and sketch simplification~\cite{SimoSerraSIGGRAPH2016}. In a typical network structure, the input gets downsampled several times to a lower dimension, then goes through a sequence of non-linear transformations, and finally gets upsampled to the desired output size. Recently, He et al. \cite{he2015residual} proposed the residual connection that uses skip layers allowing network blocks to learn only the \emph{residual} component. The use of residual block eases the training of deeper networks which improves the capability of neural network for more complex tasks.

We employ an encoder-decoder architecture with residual connections. Starting from the network design in Sketch Inversion~\cite{guccluturk2016sketchinv}, we introduce several important modifications to improve the visual quality of output and accommodate higher resolution input and more challenging image categories, such as car and bedroom. In particular, we add one more up/downsampling layer and double the number of filters in all convolutional layers between the last downsampling layer and the first upsampling step. In addition, we replace the deconvolutional layers with the bilinear upsampling step followed by two residual blocks, due to the recent finding that deconvolutional layers have the tendency to produce checkerboard artifacts commonly seen in deep generative models~\cite{odena2016deconvolution}. Overall, our architecture has around 7.8 millions learnable parameters, while the Sketch Inversion network we implemented has around 1.7 millions. See Figure~\ref{fig:network_diag} for a diagram of our architecture. 

\subsection{Objective Function}
\label{sec:objective_function}
Given pairs of training images (input, ground-truth), where the input image is derived from the ground-truth photo (synthetically generated sketches and color strokes in our case), the simplest and most common loss is the average per-pixel L2 difference between the generated image and the ground-truth, which we denote as $\mathbf{L_{p}}$. 

Previous work~\cite{guccluturk2016sketchinv} showed that adding a \emph{feature} loss to the objective function is beneficial for image generation tasks. Feature loss $\mathbf{L_{f}}$ is defined as the L2 difference in a feature space, where a feature is extracted from a certain layer of a pre-trained neural network representing high-level information of images. 

While pixel and feature losses are widely used to explicitly correlate synthesized output with input, using them alone is often not sufficient to generate diverse, realistic images. More importantly, in our problem setup, conditioning on coarse user controls leaves us with a highly ill-posed problem where the potential solution space is multimodal. Therefore, with only pixel and feature losses, the network tends to average over all plausible solutions, due to the lack of a loss which pushes for realism and diversity.

\begin{figure*}[t]
\begin{center}
\includegraphics[width=1\linewidth]{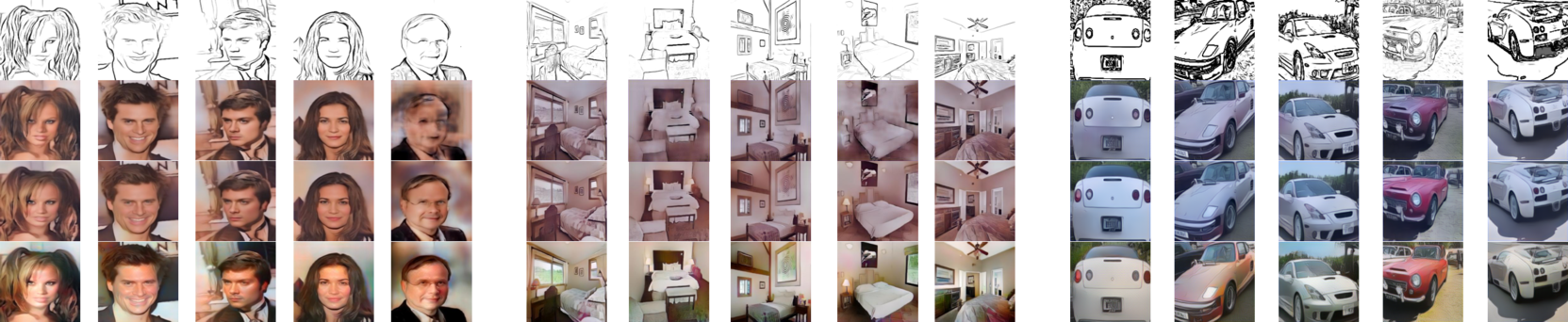}
\end{center}
   \caption{Results comparison. From top to bottom: input sketch, Sketch Inversion with content loss, our network with content loss, our network with content loss and adversarial loss}
\label{fig:adv_results}
\end{figure*}

For image categories like face, the generated results tend to have similar skin tones~\cite{guccluturk2016sketchinv}. For more complicated categories like cars and bedrooms, where the foreground and background contents can have large variety of shapes and colors, the generated results might not be visually plausible, since neutral colors are chosen by the network to minimize MSE. The second and third rows in Figure~\ref{fig:adv_results} demonstrate the problems.

To encourage more variations and vividness in generated results, we experiment with adding an adversarial loss to the objective function. Generative adversarial networks (GAN), proposed by Goodfellow et al~\cite{goodfellow2014generative}, have attracted considerable attention recently. A generative network $\mathbf{G_\theta}$ is jointly trained with a discriminative adversarial network $\mathbf{D_\phi}$, so that the discriminator
tries to distinguish between the generated images and ground-truth images, while the generator tries to fool the discriminator into thinking the generated result is real. Dosovitskiy et al~\cite{dosovitskiy2016perceptual} showed that complimenting the feature loss with an adversarial loss leads to more realistic results. The adversarial loss $\mathbf{L_{adv}}$ is defined as:
\begin{equation} \label{eq:adv}
\mathbf{L_{adv}} = -\sum{log\mathbf{D_\phi}(\mathbf{G_\theta}(x_i))}
\end{equation}
We find that adversarial loss is also beneficial for our sketch-based image synthesis problem (Figure~\ref{fig:adv_results}). With adversarial training, the network puts less emphasis on exactly reproducing ground-truth, but instead focuses on generating more realistic results with plausible color and shape deviation from ground-truth.

Adversarial training tends to be unstable, especially at the start of training when the generator does not produce anything meaningful and the discriminator can easily distinguish between real and fake. We find that using a weak discriminator $\mathbf{D_\phi}$ helps stabilize the training. We also avoided conditioning the discriminator on the input image, as this tends to increase the instability ~\cite{pathakCVPR16context}. In particular, we use a fully convolutional structure without fully connected layers and batch normalization.  Section~\ref{sec:training} introduces additional tricks for successful adversarial training.

Finally, we also add a total variation loss $\mathbf{L_{tv}}$ to encourage smoothness in the output~\cite{Johnson2016}. 

Our final objective function becomes:
\begin{equation} \label{eq:objective}
\mathbf{L} = \mathbf{w_{p}}\mathbf{L_{p}} + \mathbf{w_{f}}\mathbf{L_{f}} + \mathbf{w_{adv}}\mathbf{L_{adv}} + \mathbf{w_{tv}}\mathbf{L_{tv}}
\end{equation}

\section{Sketch-based Photo Synthesis}
\label{sec:sketch2photo}
In this section, we explore how to apply the proposed feed-forward network to hallucinate content, color and texture to reconstruct a photo based on an input sketch of arbitrary style. To train such a deep neural network, we need lots of training sketch-photo pairs. Though high quality hand-drawn sketches are readily available online, the corresponding photos based on which sketches are drawn are not. Therefore, we apply high-quality line drawing synthesis algorithms to generate synthetic sketches from photos. In order to handle real hand-drawn sketches at test time, we apply various data augmentations to the training data to improve the generality of the network. In this paper, we experiment with three image classes -- faces~\cite{liu2015faceattributes}, cars, and bedrooms~\cite{Yu2015}. We believe the proposed framework can generalize well to other categories given similar amounts of training data and training time.

\begin{figure}[b]
\begin{center}
\includegraphics[width=1\linewidth]{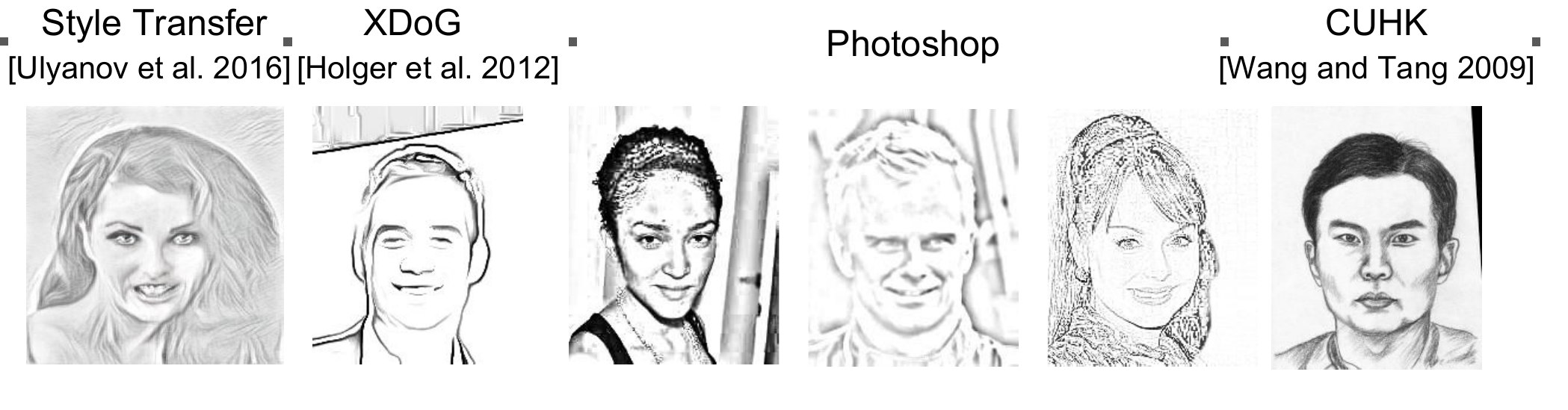}
\end{center}
   \caption{We generate synthetic sketches from photos using five different algorithms. We also include and augment a small set of hand-drawn sketch-photo pairs to help generalize the network to handle real hand-drawn sketch inputs.}
\label{fig:multiple_styles}
\end{figure}

\subsection{Generation of Training Sketches}
\label{sec:data_aug}
For each image category -- face, car, or bedroom -- we apply the boundary detection filter XDoG~\cite{winnemoller2012xdog} on 200k photos to generate the corresponding synthetic sketches. The input (and output) resolution to our network during the training phase is 128x128.

To make the network invariant to the exact locations of the objects, we randomly crop both the input and the ground-truth images. For the face and bedroom categories, we first resize the images to 256x256 before randomly cropping them to 128x128. For the car category, we scale the images to 170x170 before cropping, since most cars already occupy large image areas, enlarging them too much means losing the global spatial arrangement and the contexts around the cars. 

In addition to randomly cropping an image and its corresponding sketch, we also randomly adjust the brightness level of the sketch to get different levels of details from the same sketch (i.e. some sketch lines will disappear with higher brightness level). Finally, we also randomly cut off some lines in the sketch, by overlaying a random number of white strokes (the background color of sketch input) on top of the sketch. We randomize the length, width and locations of the white strokes.

\begin{figure*}[t]
\begin{center}
\includegraphics[width=\linewidth]{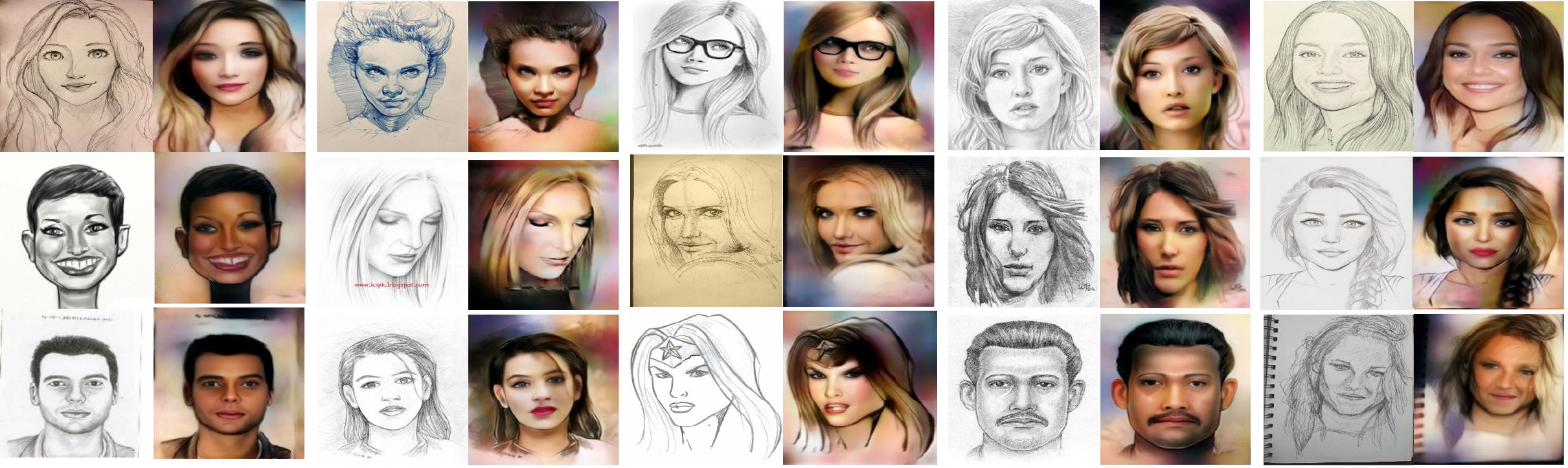}
\end{center}
   \caption{Sketch-based photo synthesis of hand-drawn test sketches. Despite the diverse sketch styles, our network usually produces high quality, diverse results. Notice the skin tone and hair color variations in the output. Some outputs are non-photorealistic because they are being somewhat faithful to caricatured input sketches. Unfortunately, some results have unrealistically high low-frequency contrast and appear unnaturally lit.}
\label{fig:results_styles}
\end{figure*}

\subsection{Network Generalization}
\label{sec:generalization}
Real hand-drawn sketches exhibit a large variety of styles, from abstract pen-and-ink illustrations to elaborate pencil-like drawings with shading. The characteristics of the hand-drawn sketches might be very different from the synthetic sketches we generated algorithmically. Even with the various augmentations, random cropping, random brightness adjustment and random cut-off, the trained network might still overfit to that particular style of sketches. To improve the network generality, we further augment the training data by adding multiple styles of sketches.

For the face category, we obtain 20k additional images and for each image we randomly choose one of the following four algorithms to synthesize a corresponding sketch. See example sketches in Figure~\ref{fig:multiple_styles}.

\begin{itemize}
\item \textbf{StyleNet}~\cite{gatys2016image} We apply neural network-based  style transfer algorithm to transfer the texture style of a pencil drawing to the ground-truth photo.
\item \textbf{Photoshop filters}~\cite{photoshopFilter2} Applying Photoshop's 'photocopy' effect to ground-truth images, we can generate two different versions of sketches with different levels of details and stroke darkness. 
\item \textbf{Gaussian blur on inverse image}~\cite{photoshopFilter1} Using Photoshop, we can also synthesize another sketch style by performing Gaussian blur on an inverse (grayscale) image in Photoshop color dodge mode. This creates detailed line drawings with very little shading. 
\item \textbf{CUHK} Finally, we add the CUHK dataset, which contains 188 hand-drawn portrait sketches and their corresponding photos \cite{wang2009CUHK}. To give higher weights to the high quality hand-drawn sketches, we apply mirroring and varying degrees of rotation to the sketches and end up with 1869 images in total.
\end{itemize}

At this point, we have 21848 images of 6 different sketch styles. 
Pre-trained on the 200k sketches of the XDoG style, the network is fine-tuned using the 20k multi-style sketches. We use the same parameter settings as before and train the network on these additional data for 5 epochs.

\begin{figure}[t]
\begin{center}
\includegraphics[width=0.8\linewidth]{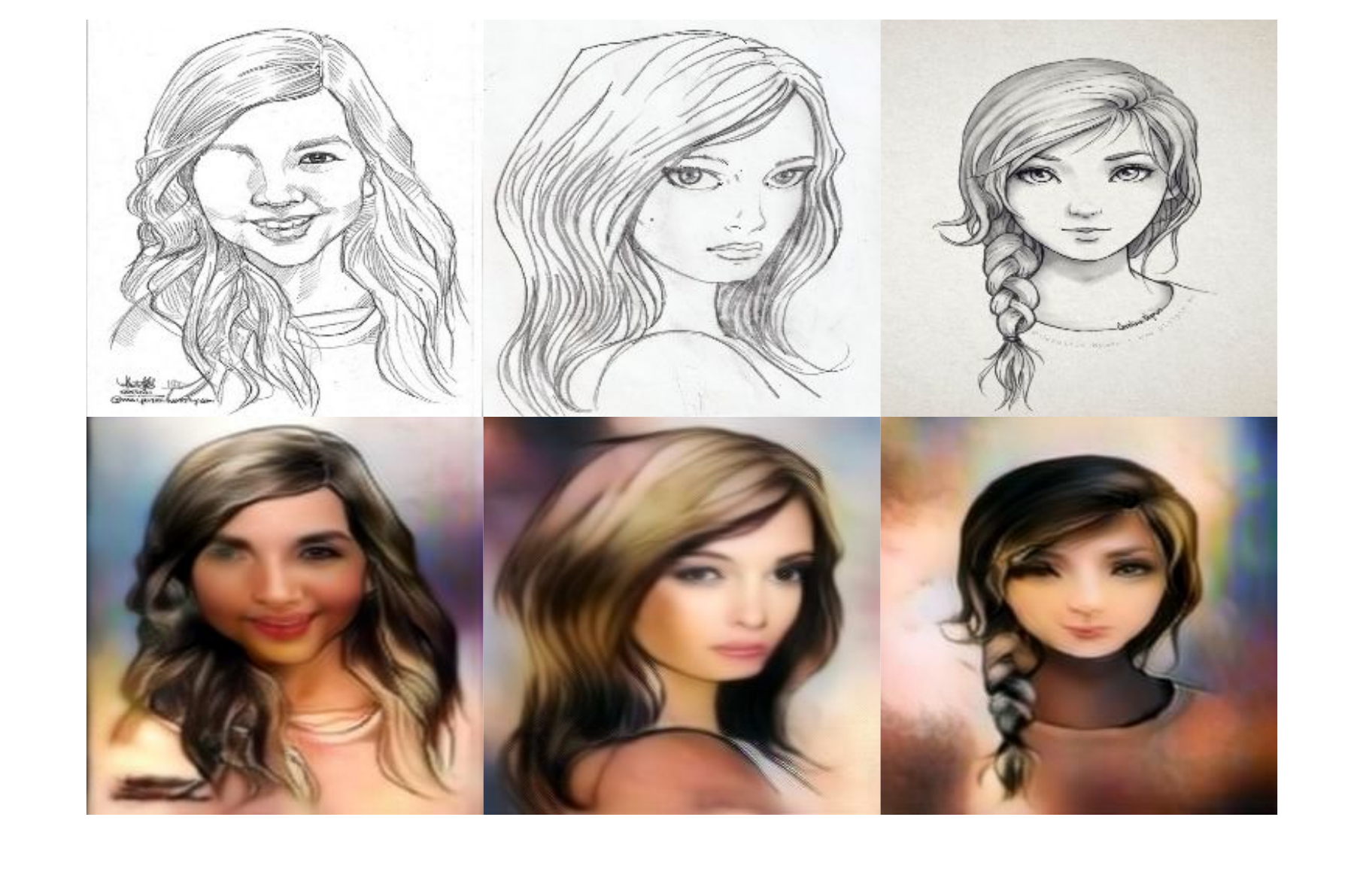}
\end{center}
   \caption{Interesting network behavior. Left: the network hallucinates the missing eye. Middle: the adversarial training reduces the size of the eyes to more realistic proportions. Right: The network curves the lips upward changing the overall expression to a subtle smile.}
\label{fig:weird_stuff_gan_do}
\end{figure}

\subsection{Results and Discussions}
\label{sec:sketchResults}

For comparison purposes, we implemented the Sketch Inversion architecture as described in~\cite{guccluturk2016sketchinv}. We trained both the Sketch Inversion network and our deeper network using the same training data and parameter settings.
Figure~\ref{fig:adv_results} shows side-by-side comparisons of the results generated by Sketch Inversion (second row), our deeper network trained without (third row) and with adversarial loss (fourth row) on three different image categories. Compared to Sketch Inversion, our deeper network even without adversarial loss produces sharper results on complex bedroom scenes and performs better at hallucinating missing details (shapes of eyes and eyebrows) given simplified sketches with few lines. 
With adversarial loss, our network is encouraged to generate images with sharper edges, higher contrast and more realistic color and lighting. As discussed in Section~\ref{sec:objective_function}, adversarial loss helps the network generate more diversified results, avoiding always producing similar skin tones and hair colors for portraits and dull and unrealistic colors for the bedrooms and cars. Figure~\ref{fig:results_styles} shows diverse hair colors and skin tones in the result.

Among the three image categories, bedroom is arguably most challenging, since each bedroom scene can contain multiple object categories. The fact that our current network handles it successfully with only 200K training data leads us to believe the possibility of training a general sketch-to-photo network across several image categories using an even deeper network. 

\begin{figure*}
\begin{center}
\includegraphics[width=1\linewidth]{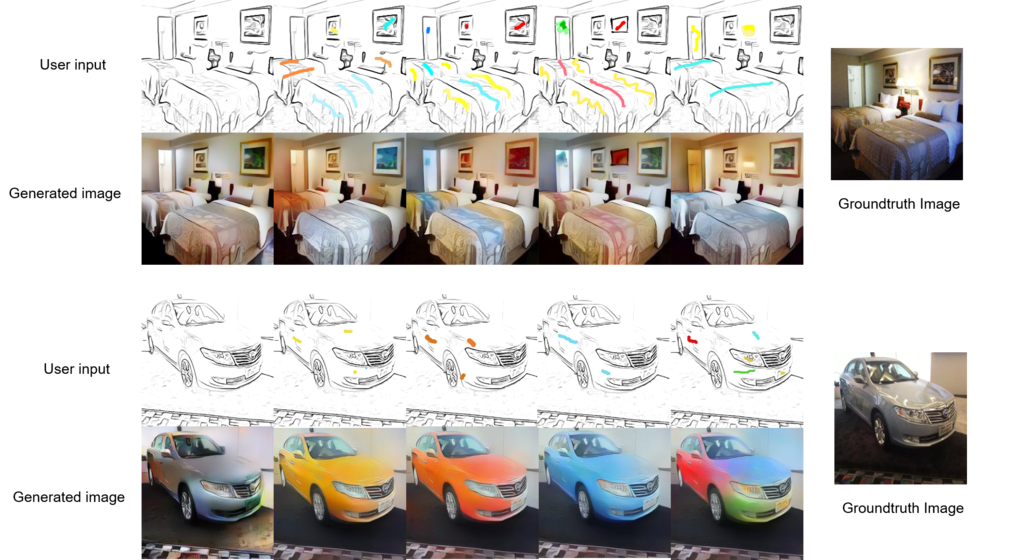}
\end{center}
   \caption{Guided sketch colorization results on held out test sketches. Without color strokes, our network produces synthesis results that closely follow the input sketch (leftmost column). With color strokes, our network can adapt the synthesis results to satisfy different color constraints. Note that the fourth and fifth bedroom sketches are edited to add a framed picture on the wall and a lamp using only simple user edits.}
\label{fig:stroke_testing}
\end{figure*}

After training with multiple sketch styles and various data augmentations (Section~\ref{sec:generalization}), our network generates much more realistic results given arbitrary hand-drawn sketches as input. 
Figure~\ref{fig:results_styles} shows reconstruction results based on sketches found by Google search. Note that the sketches are drawn with diverse styles, some detailed and realistic, some abstract and simplified. The results show that our network generalizes well to arbitrary hand-drawn sketches and is robust to the variations in head pose, background colors, and textures. 
Data augmentation such as random cropping and cutoff also helps our network hallucinate missing details. Figure~\ref{fig:weird_stuff_gan_do} (left) shows that the network can fill in the missing eye to some extent. However, generating missing object parts is a challenge itself, therefore we consider it beyond the scope of this paper.

The network trained with adversarial loss has an interesting behavior. When applying it to cartoonish or unprofessional sketches with intentional or unintentional exaggeration of facial features, the network tends to `realistify' or beautify the input sketch to generate result more photo-like at the cost of not strictly following the sketch constraints.
For example, eyes that are inhumanly large will get reduced to a realistic size or faces with weird shapes will be smoothed and `beautified' (see Figure~\ref{fig:weird_stuff_gan_do}). To produce realistic results, the network learns not to blindly trust the sketch input, but resorts to its understanding about the natural image manifold acquired during the adversarial training.

\section{User-guided Colorization}
\label{sec:colorization}
The previous section focuses on using a gray-scale sketch to guide the generation of a color photos. The lack of color information in the input causes the problem to be under-determined, since one sketch can correspond to photos colored in many different ways.

Although the use of adversarial loss constrains the output to lie on an approximated manifold of natural images and therefore limits the color choices, it is still up to the generator (and the discriminator) to choose a specific color. 

In this section, we explore how to allow users to directly control the colors in the output. To do that, we need to modify the input to the network to include rough color information during training (Section~\ref{sec:color_stroke}). We investigated adding color controls in two applications, guided sketch colorization (Section~\ref{sec:sketchColorization}) and guided image colorization (Section~\ref{sec:grey_colorization}).

\subsection{Generation of Training Color Strokes}
\label{sec:color_stroke}
One of the most intuitive ways to control the outcome of colorization is to `scribble' some color strokes to indicate the preferred color in a region. 

To train a network to recognize these control signals at test time, we need to synthesize color strokes for the training data. We generate synthetic strokes based on the colors in the ground-truth image. 

To emulate arbitrary user behaviors, we blur the ground-truth image and sample a random number of color strokes of random length and thickness at random locations. We pick the ground-truth  pixel color at the stroke starting point as the stroke color and continue to grow the stroke until the maximum length is reached. 

When growing a stroke, if the difference between the current pixel color and the stroke color exceeds a certain threshold, we restart the stroke with a new color sampled at the current pixel. By randomizing various stroke parameters, we are able to synthesize color strokes similar to what human would draw during test time.

\subsection{Guided Sketch Colorization}
\label{sec:sketchColorization}

The goal here is to add color control to our sketch-based image synthesis pipeline. Our previous objective function still holds: we want the output to have the same content as the input (pixel and feature loss), and appear realistic (adversarial loss). Pixel loss is essential here as it forces the network to be more precise with color by paying more attention to the color strokes. We modify the training data by placing color strokes on top of the input sketches. 
We then train the network as before using a parameter setting that emphasizes content loss and de-emphasizes adversarial loss, so that the results better satisfy color constraints (Section~\ref{sec:adversarialTraining}). 

Figure~\ref{fig:stroke_testing} shows the results of reconstructing bedroom and car scenes based on an input sketch and color strokes. Note that the colors of the strokes deviate a lot from the colors in the ground-truth image, nevertheless, the network is able to propagate the input color to the relevant regions respecting object boundaries. In the bedroom scene (two rightmost columns), based on the crude outline of a picture frame and a yellow lamp, our network successfully generates plausible details in the results. See more results in the supplementary material.

\subsection{Guided Image Colorization}
\label{sec:grey_colorization}

\begin{figure}
\begin{center}
\includegraphics[width=\linewidth]{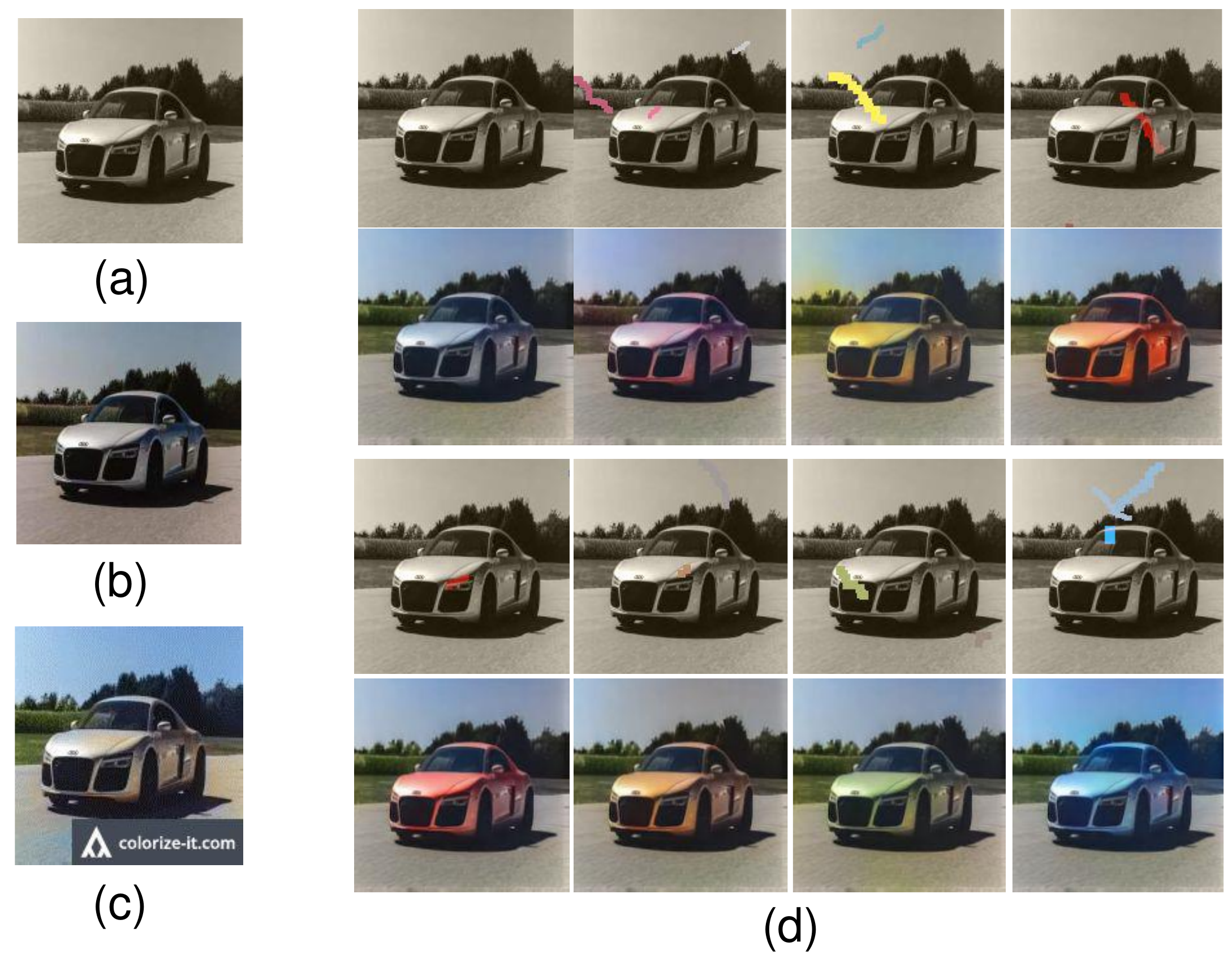}
\end{center}
\caption{Guided Image Colorization: a) grayscale input, b) original color image, c) deep colorization result~\cite{zhang2016colorful}, d) First and third rows: color strokes overlaid on top of the grayscale input (zoom in to see the color strokes). Second and fourth rows: colorization results. }
\label{fig:grey_colorization}
\end{figure}

Recent work~\cite{IizukaSIGGRAPH2016letbecolor,zhang2016colorful} explores training deep neural network models for the image colorization tasks. However, the selection of colors in the output is entirely up to the network. In this section, we investigate using color strokes (Section~\ref{sec:color_stroke}) to guide the colorization process. We generate training data by extracting a one-channel grayscale image from the ground-truth photo and combining it with the three-channel image containing color strokes. 

Figure~\ref{fig:grey_colorization} shows various colorization results on a car image. Given a gray-scale image, our system synthesizes realistic looking cars based on strokes drawn with different colors at random locations. Note that most strokes are placed on the body of the car and therefore do not influence the colorization of the other regions. Due to adversarial training, the sky is colored blue and the trees are colored green, regardless of the colors of the foreground object. Internally, the network learns to recognize semantic contents and therefore can put right colors in the relevant regions while at the same time satisfying user constraints wherever possible. 

See more results in the supplementary material.

\section{Applications}
\label{sec:apps}
We believe being able to control the generated output using sketch and color allows for many useful applications, especially in the artistic domain. 
\subsection{Interactive Image Generation Tools}

\begin{figure}
\begin{center}
\includegraphics[width=1\linewidth]{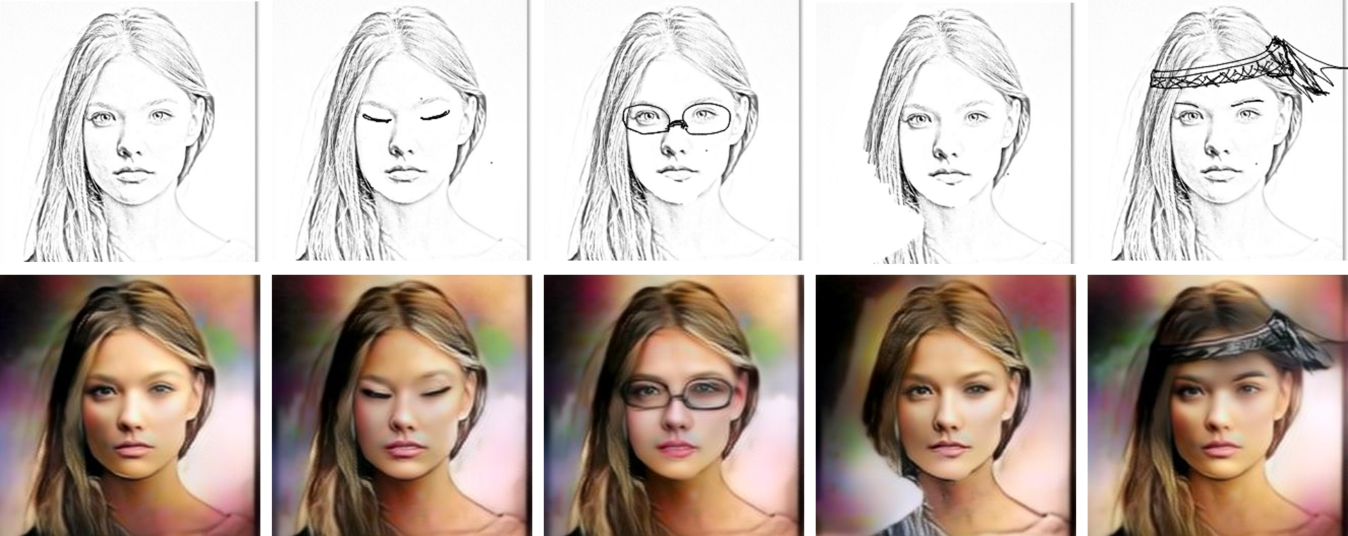}
\end{center}
   \caption{Interactive image generation and editing. The user can incrementally modify the sketch to change the eyes, hair, and head decorations. }
\label{fig:image_editing}
\end{figure}

Given an input image of resolution 256x256, our network takes about 20ms to transform it to a photo-like result. 
The real-time performance enables instant visual feedback after incremental edits in image generation and editing applications. 

Using sketch and color strokes to enforce fine-grained control is useful for several design applications. For example, an interior designer can quickly sketch out rough shapes of the objects, specify colors in various regions and let our system fill in missing details and textures to generate a plausible bedroom scene. After seeing the result, the designer can interactively modify the shapes and colors of the objects and receive instant visual feedback. Figure~\ref{fig:stroke_testing} illustrates the potential design workflow. Similarly, a car designer can follow similar workflow to design new cars and test out the looks in different background settings. 

Our portrait synthesis system provides a tool for artists to design virtual characters (see Figure~\ref{fig:image_editing}). Based on the initial design, one can change the shape of eyes and/or hairstyle, add glasses and/or head decorations, etc. In addition to design, portrait reconstruction technology is useful for forensic purposes, for example the law enforcement department can use it to help identify suspects~\cite{guccluturk2016sketchinv}.

\subsection{Sketch and Color Guided Visual Search}
\begin{figure}
\begin{center}
\includegraphics[width=0.7\linewidth]{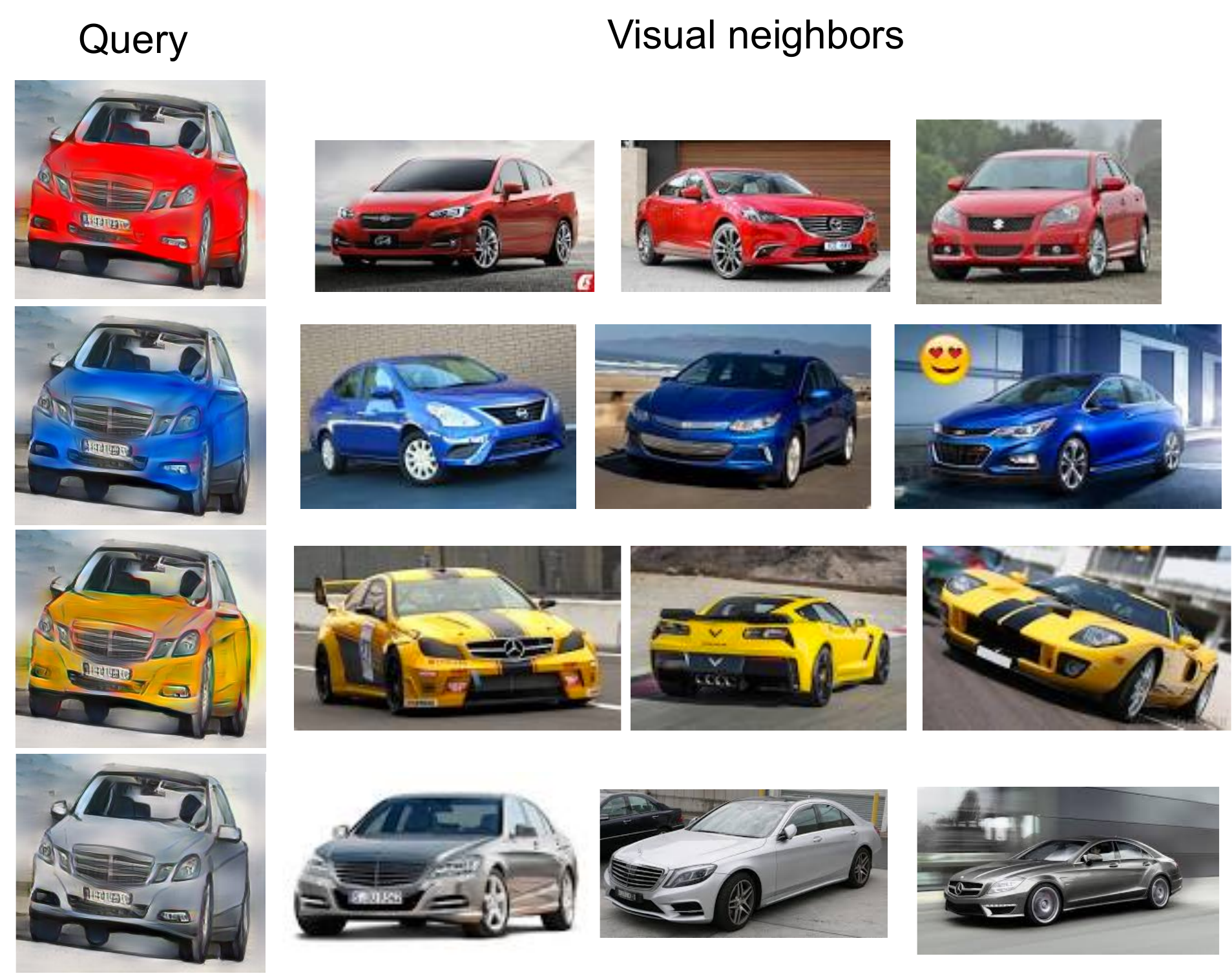}
\end{center}
   \caption{Given the reconstructed images (leftmost column), Google's visual search engine retrieves visually similar photos.}
\label{fig:guidedsearch}
\end{figure}
Our image generation tools provide flexible ways to perform visual search. With a target scene in mind, one can sketch out the object boundaries and color constraints, based on which our network can reconstruct a plausible arrangement. The reconstructed image can then be used in a typical visual search tool to identify high-res images with similar contents (see Figure~\ref{fig:guidedsearch}). 

\section{Network Training Details}
\label{sec:training}
With the unpredictable nature of adversarial training, we find it helpful to separate the training into two stages. 

\subsection{Optimizing for Content Loss}
In the first stage, we set the adversarial weight $\mathbf{w_{adv}}$ from equation~\ref{eq:objective} to 0 and let the network focus on minimizing the content loss which is a combination of pixel and feature loss. To enforce a fine-grained control using the input sketch, we choose the ReLU2-2 layer of the VGG-19 net~\cite{simonyan2014very} to compute the feature loss, since higher level feature representations tend to encourage the network to ignore important details such as the exact locations of the pupils.

We set the weights of pixel loss and feature loss $\mathbf{w_p}$, $\mathbf{w_f}$ to 1, and the weight of TV loss $\mathbf{w_{tv}}$ to 1e-5. We train the network for around 3 epochs using a batch size of 32 before moving on to the second stage of the training.

\subsection{Adding Adversarial Loss}
\label{sec:adversarialTraining}
Given the network pretrained for content loss, we fine tune it with different loss settings for different applications. 
For photo reconstruction from gray-scale sketches (Section~\ref{sec:sketch2photo}), we turn off the pixel loss, keep the feature loss and add the adversarial loss with the following weight setting, $\mathbf{w_f}=1, \mathbf{w_p}=0, \mathbf{w_{tv}}=0, \mathbf{w_{adv}}\approx1e8$. 
For colorization applications (Section~\ref{sec:colorization}), we emphasize the feature and pixel loss and de-emphasize the adversarial loss, so that the output better follows the color controls, $\mathbf{w_f}=10, \mathbf{w_p}=1, \mathbf{w_{tv}}=0, \mathbf{w_{adv}}\approx1e5$.

We train the adversarial discriminator alongside our generative network for three epochs using a learning rate between 1e-5 and 1e-6. 

\section{Conclusion and Future Work}
In this paper, we propose a deep generative framework that enables two types of user controls to guide the result generation -- using sketch to guide high-level visual structure and using sparse color strokes to control object color pattern. 

Despite the promising results, our current system  suffers from several limitations. 
First, we sometimes observe blurry boundaries between object parts or regions of different colors which diminish the overall realism of the results. 

Figure~\ref{fig:stroke_testing} shows the color 
leaking problem on the car results, where the color of the car's hood leaks into the background.
Second, our system struggles between strictly following color/sketch controls and minimizing adversarial loss. In other words, adversarial loss prohibits the generated images from taking uncommon colors and shapes. If the user specifies a rare color, for example, purple for car, red for trees, our network will map it to a different color deemed more realistic by the adversarial loss. 
Third, the network sees objects of similar scale during training, and would expect to see the same scale at testing. As future work, we can add multi-scale support to the network by randomizing the ratio between the cropping size and the image size during training. 

Going forward, we would like to investigate how to further improve the visual results by encouraging sharp color boundaries and finding systematic ways to deal with rare control signals.

\section*{Acknowledgments}
We thank Yijun Li for assistance with generation of synthetic training sketches from ~\cite{gatys2016image}. This work is supported by a Royal Thai Government Scholarship to Patsorn Sangkloy, NSF CAREER award 1149853 to James Hays, and NSF award 1561968. 

{\small
\bibliographystyle{ieee}
\bibliography{egbib}
}



\begin{figure*}[ht]
\centering
\includegraphics[width=.95\linewidth]{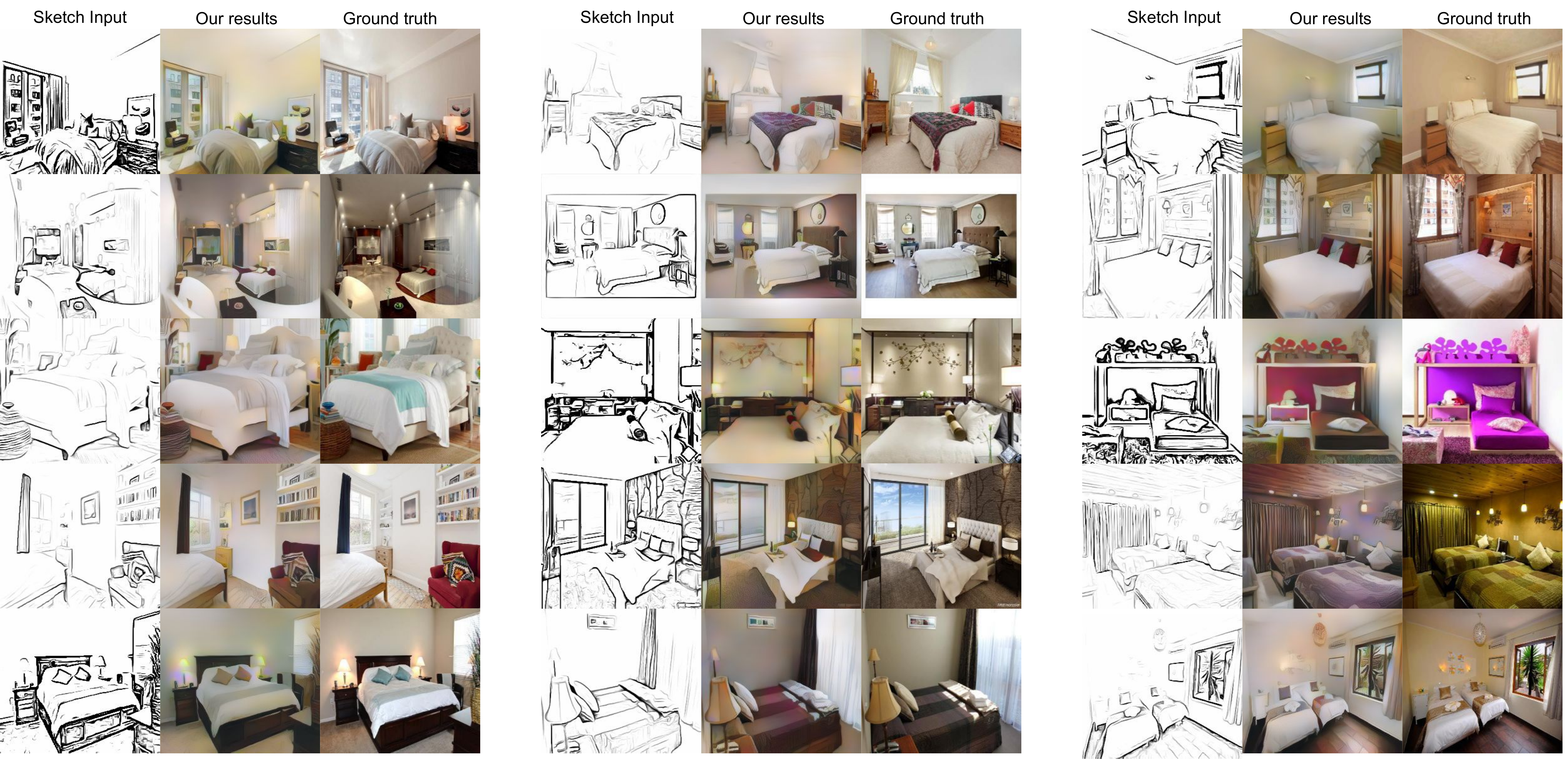}
\caption{Additional results on held out bedroom sketches}
\end{figure*}


\begin{figure*}[ht]
\centering
\includegraphics[width=1\linewidth]{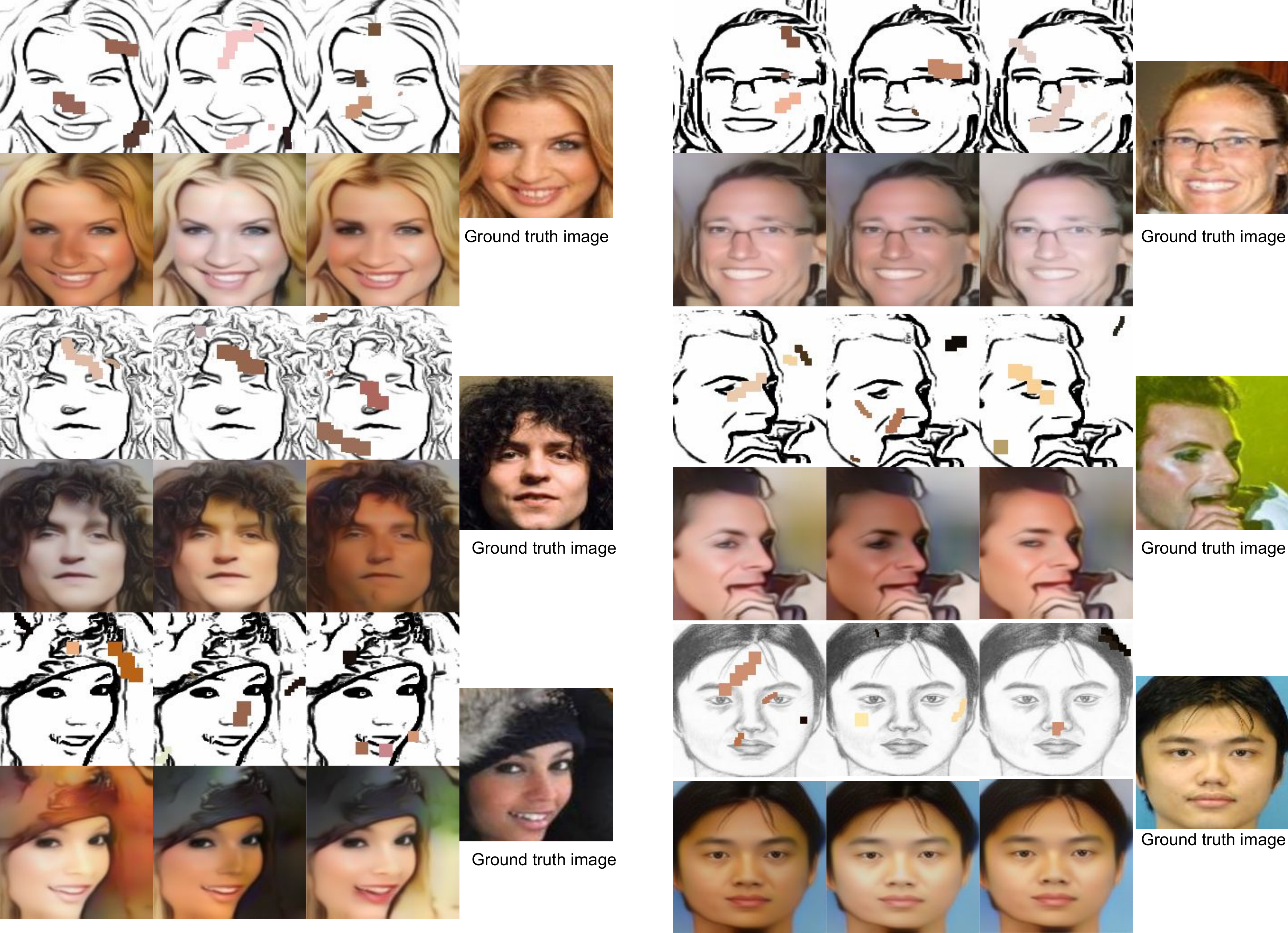}
\caption{Additional results on held out face sketches of multiple styles with random color strokes. The color strokes are generated by sampling curves from random face images.}
\end{figure*}

\begin{figure*}[ht]
\centering
\includegraphics[width=1\linewidth]{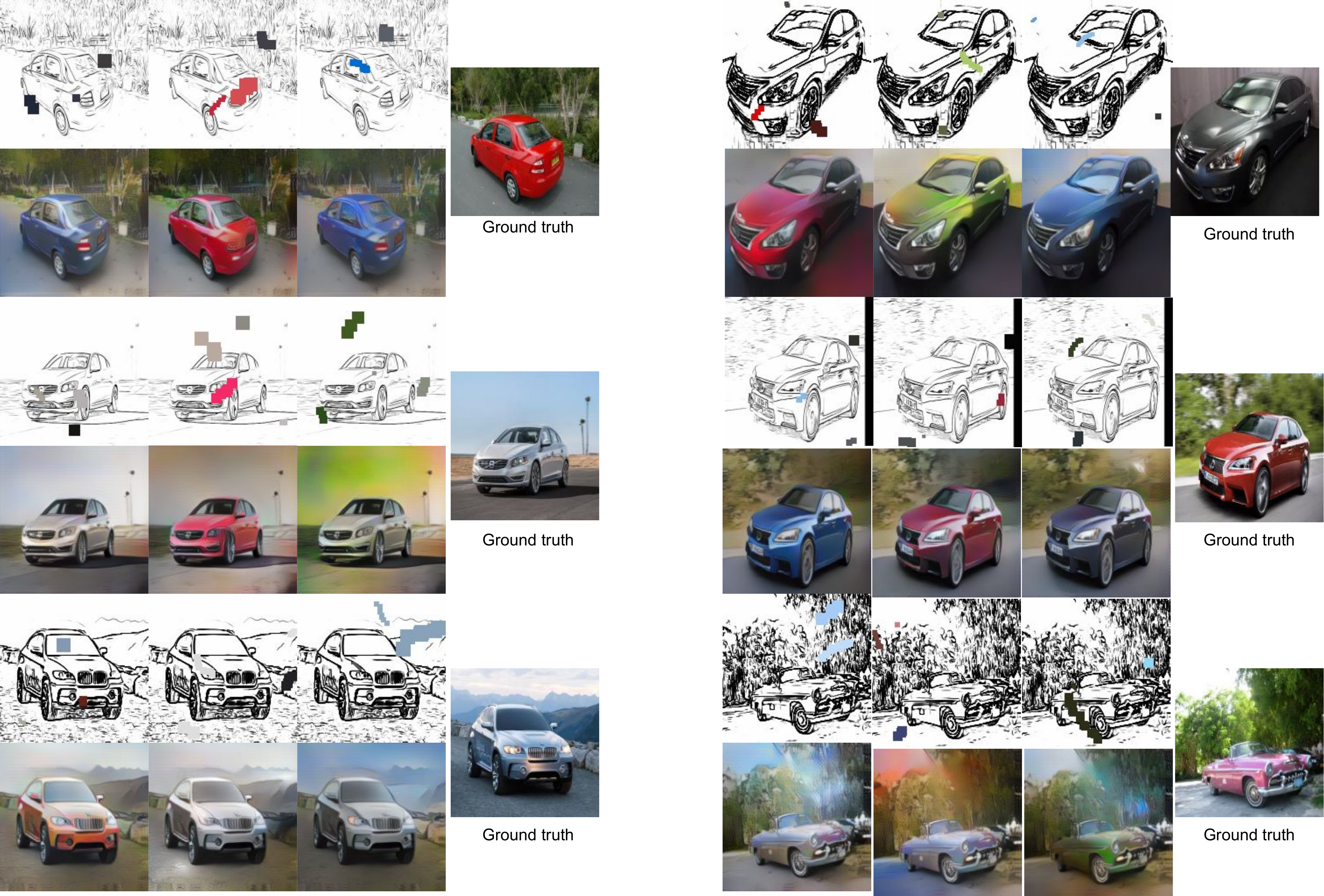}
\caption{Additional results on held out car sketches with random color strokes. The color strokes are generated by sampling curves from random car images.}
\end{figure*}


\begin{figure*}[ht]
\centering
\includegraphics[width=1\linewidth]{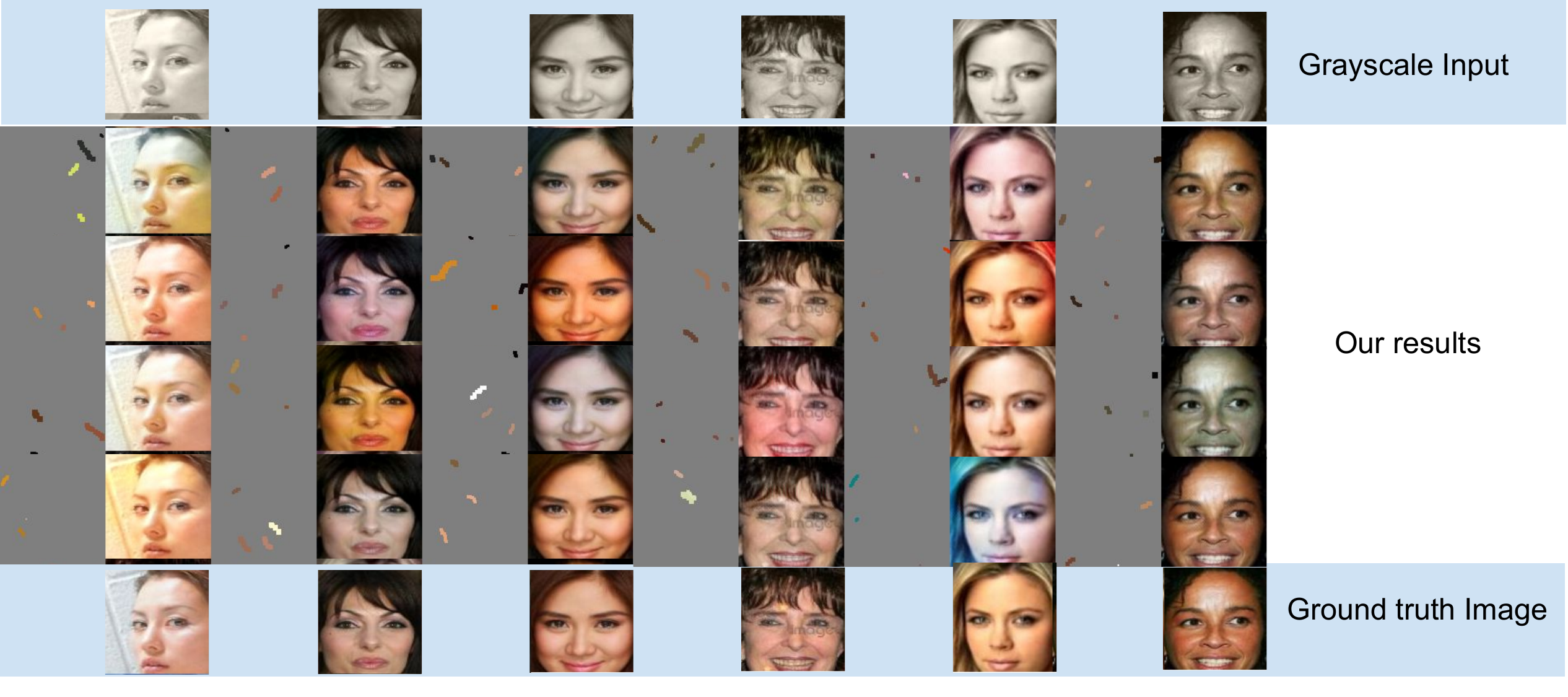}
\caption{Additional face colorization results with random color suggestions. The color strokes are sampled from random curves in other face images.}
\end{figure*}


\end{document}